\documentclass{llncs}
\usepackage{makeidx}  
\usepackage{graphicx}

\usepackage{placeins}

\usepackage[latin1]{inputenc}

\usepackage{amsmath}
\usepackage{amsfonts}
\usepackage{amssymb}

\usepackage{mathrsfs}
\usepackage{scalerel}

\usepackage{url}    
\newcommand{\keywords}[1]{\par\addvspace\baselineskip
\noindent\keywordname\enspace\ignorespaces#1}

\newcommand\scale[2]{\vstretch{#1}{\hstretch{#1}{#2}}}
\newcommand{\LIPplus}{\mathbin{\ooalign{$\bigtriangleup$\crcr\hidewidth
  \raise.14em\hbox{$\scale{0.7}{\scriptscriptstyle+}$}\hidewidth}}}
\newcommand{\LIPminus}{\mathbin{\ooalign{$\bigtriangleup$\crcr\hidewidth
  \raise.14em\hbox{$\scale{0.7}{\scriptscriptstyle-}$}\hidewidth}}}
\newcommand{\LIPtimes}{\mathbin{  \ooalign{$\bigtriangleup$\crcr\hidewidth
  \raise.14em\hbox{$\scale{0.7}{\scriptscriptstyle\times}$}\hidewidth}}}

\newcommand{\Real}{\mathbb R}
\newcommand{\Tcurv}{\mathcal{T}}

\newcommand{\la}{\lambda}

\newcommand{\I}{\mathcal{I}}
\newcommand{\Ieq}{\I^{\LIPtimes}}
\newcommand{\ft}{f^{\LIPtimes}}
\newcommand{\gt}{g^{\LIPtimes}}
\newcommand{\htt}{h^{\LIPtimes}}
\newcommand{\Lscr}{\mathscr{L}}

\usepackage{verbatim}

\DeclareFontFamily{U}{mathx}{\hyphenchar\font45}
\DeclareFontShape{U}{mathx}{m}{n}{
      <5> <6> <7> <8> <9> <10>
      <10.95> <12> <14.4> <17.28> <20.74> <24.88>
      mathx10
      }{}
\DeclareSymbolFont{mathx}{U}{mathx}{m}{n}
\DeclareFontSubstitution{U}{mathx}{m}{n}
\DeclareMathAccent{\widecheck}{0}{mathx}{"71}

\begin{document}
%

\mainmatter              
\title{Double-sided probing by map of Asplund's distances using Logarithmic Image Processing in the framework of Mathematical Morphology}
\titlerunning{Map of Asplund's metrics \& Mathematical Morphology}  
%
\author{Guillaume Noyel\inst{1} \and Michel Jourlin\inst{1,2}}
\authorrunning{Guillaume Noyel et al.} 
%
\tocauthor{Guillaume Noyel, and Michel Jourlin}
\institute{International Prevention Research Institute,	95 cours Lafayette,	69006 Lyon, France
\and Lab. H. Curien, UMR CNRS 5516,	18 rue Pr. B. Lauras,	42000 St-Etienne, France\\
\url{www.i-pri.org}}

\maketitle              

\begin{abstract}
We establish the link between Mathematical Morphology and the map of Asplund's distances between a probe and a grey scale function, using the Logarithmic Image Processing scalar multiplication. We demonstrate that the map is the logarithm of the ratio between a dilation and an erosion of the function by a structuring function: the probe. The dilations and erosions are mappings from the lattice of the images into the lattice of the positive functions. Using a flat structuring element, the expression of the map of Asplund's distances can be simplified with a dilation and an erosion of the image; these mappings stays in the lattice of the images. We illustrate our approach by an example of pattern matching with a non-flat structuring function.

\keywords{Map of Asplund's distances, Mathematical Morphology, dilation, erosion, Logarithmic Image Processing, Asplund's metric, double-sided probing, pattern recognition}
\end{abstract}
%

%
%

\section{Introduction}

Asplund's metric is a useful method of pattern matching based on a double-sided probing, i.e. a probing by a greatest lower bound probe and a least upper bound probe. It was originally defined for binary shapes, or sets \cite{Asplund1960,Grunbaum1963}, by using the smallest homothetic shape (probe) containing the shape to be analysed and the greatest homothetic probe contained by the shape. Jourlin et al. \cite{Jourlin2012,Jourlin2014} have extended this metric to functions and to grey-level images in the framework of the Logarithmic Image Processing (LIP) \cite{Jourlin1988,Jourlin2001} using a multiplicative or an additive LIP law \cite{Jourlin2016_chap3}. Then, Asplund's metric has been extended to colour and multivariate images by a marginal approach in \cite{Noyel2015,Jourlin2016_chap3} or by a spatio-colour (i.e. vectorial \cite{Noyel2007,Noyel2014}) approach in \cite{Noyel2016}.

Other approaches of double-sided probing have been previously defined in the framework of Mathematical Morphology \cite{Matheron1967,Serra1982}. The well-known hit-or-miss transform \cite{Serra1982} allows to extract all the pixels such that the first set of a structuring element fits the object while the second set misses it (i.e. fits its background). An extension based on two operations of dilation (for grey level images) has been proposed in \cite{Khosravi1996}. It consists of a unique structuring element, which is used in the two dilations in order to match the signal from above and from below.

Banon et al. \cite{Banon1997} use two structuring elements obtained by two translations of a unique template along the grey level axis. They use an erosion and an anti-dilation to count the pixels whose values are in between the two structuring elements.

Odone et al. \cite{Odone2001} use an approach inspired by the computation of the Hausdorff distance. They consider a grey level image as a tridimensional (3D) graph. They dilate by a 3D ball a template in order to compute a 3D ``interval''. Then, for any point of the image, they translate vertically the ``interval'' in order to contain the maximum number of points of the function and they count this number.

Barat et al. \cite{Barat2010} present a unified framework for these last three methods. They show that they correspond to a neighbourhood of functions (i.e. a tolerance tube) with a different metric for each method. Their topological approach is named virtual double-sided image probing (VDIP) and they defined it as a difference between a grey-scale dilation and an erosion. For pattern matching, only the patterns which are in the tolerance tube are selected. It is a metric defined on the equivalence class of functions according to an additive grey level shift.

In \cite{Jourlin2012}, Jourlin et al. have introduced the logarithmic homothetic defined according to the LIP multiplication. This makes a compensation of the lighting variation due to a multiplicative effect, i.e. a thickening or a thinning of the object crossed by the light.

In the current paper, the important novelty introduced is the link between the map of Asplunds' metrics defined in the LIP multiplicative framework and the operations of Mathematical Morphology. We will show that the map of Asplunds' distances in the LIP multiplicative framework is the logarithm of the ratio between a dilation and an erosion.

This gives access to many other notions well defined in the corpus of Mathematical Morphology.

The paper is organised as follows: 1) a reminder of the main notions (LIP, Asplund's metrics, fundamental operations and framework of Mathematical Morphology), 2) the demonstration of the link between the map of Asplund's distances and Mathematical Morphology for flat structuring element (se) and for non-flat ones and 3) an illustration of pattern matching with Asplund's metric.

%
%

\section{Prerequisites}

In the current section, we remind the different mathematical notions and frameworks to be used: LIP model, Asplund's metric and the basis of Mathematical Morphology.

\subsection{Logarithmic Image Processing (LIP)}

The LIP model, created by Jourlin et al. \cite{Jourlin1988,Jourlin1995,Jourlin2001,Jourlin2016_chap1}, is a mathematical framework for image processing based on the physical law of transmittance. It is perfectly suited to process images acquired with transmitted light (when the object is located between the source and the sensor) but also with reflected light, due to the consistency of the model with human vision \cite{Brailean1991}.
The mathematical operations performed using the LIP model are consistent with the physical principles of image formation. Therefore the values of an image defined in $[0, M[$ stay in this bounded domain. For 8 bits images $M=256$ and the 256 grey levels are in the range of integers $[0, ..., 255]$. 

A grey scale image $f$ is a function defined on a domain $D \subset \Real^N$ with values in $\Tcurv = \mathcal[0,M[$, $M \in \Real$. $f$ is a member of the space  $\I = \Tcurv^D$.

Due to the link with the transmittance law: $T_f = 1-f/M$, the grey-scale is inverted in the LIP framework, $0$ corresponds to the white extremity of the grey scale, when no obstacle is located between the source and the sensor, while the other extremity $M$ corresponds to the black value, when the source cannot be transmitted through the obstacle.

In the LIP sense, the addition of two images corresponds to the superposition of two obstacles (objects) generating $f$ and $g$:
\begin{equation}
  f \LIPplus g = f + g - \frac{f.g}{M}
  \label{eq:LIPplus}
\end{equation}

From this law, we deduce the LIP multiplication of $f$ by a scalar $\la \in \Real$:
\begin{equation}
  \lambda \LIPtimes f = M - M \left( 1 - \frac{f}{M} \right)^{\lambda}
  \label{eq:LIPtimes}
\end{equation}

It corresponds to a thickness change  of the observed object in the ratio $\lambda$. If $\lambda>1$, the thickness is increased and the image becomes darker than $f$, while if $\lambda \in [0,1[$, the thickness is decreased and the image becomes brighter than $f$.

The LIP laws satisfy strong mathematical properties. Let $\mathcal{F}(D,[-\infty,M[)$ be the set of functions defined on $D$ with values in $]-\infty,M[$. We equipped it with the two logarithmic laws and $(\mathcal{F}(D,[-\infty,M[),\LIPplus,\LIPtimes)$ becomes a real vector space. $(\I,\LIPplus,\LIPtimes)$ is the positive cone of this vector space \cite{Jourlin2001}.

There exists a colour version of the LIP model \cite{Jourlin2011}.

The LIP framework has been successfully applied to numerous problems for industry, medical applications, digital photography, etc. It gives access to new notions of contrast and metrics which take into account the variation of light, for example the Asplund's metric for functions.

\subsection{Asplund's metric for functions using the LIP multiplicative law}

Let us remind the novel notion of Asplund's metric defined in \cite{Jourlin2012,Jourlin2014} for functions in place of sets. It consists of using the logarithmic homothetic $\la \LIPtimes f$.

Let $\Tcurv^* = ]0,M[$ and the space of positive images be $\I^*={\Tcurv^*}^D$.

\begin{definition}{\textbf{Asplund's metric}}
Given two images $f$, $g \in \I^*$, $g$ is chosen as the probing function for example, and we define the two
numbers: $\lambda = \inf \left\{\alpha, f \leq \alpha \LIPtimes g \right\}$ and $\mu = \sup \left\{\beta, \beta \LIPtimes g \leq f\right\}$. 
The corresponding ``functional Asplund's metric'' $d_{As}^{\LIPtimes}$ (with the LIP multiplication) is:
\begin{equation}\label{eq:das}
	d_{As}^{\LIPtimes}(f,g) = \ln \left( \lambda / \mu \right)
\end{equation}
\label{def:asplund_metric}
\end{definition}
From a mathematical point of view \cite{Jourlin2016_chap3}, $d_{As}^{\LIPtimes}$ is a metric if the images $f, g \in \I^*$  are replaced by their equivalence classes $f^{\LIPtimes}=\{g / \exists k >0,  k \LIPtimes g = f  \}$ and $g^{\LIPtimes}$. 

The relation $(\exists k >0,  k \LIPtimes g = f)$ is clearly an equivalence relation written $fRg$, because it satisfies the three properties: i) reflexivity $\forall f \in \I, fRf$, ii) symmetry $\forall (f,g) \in \I^2$, $fRg \Leftrightarrow gRf$ and iii) transitivity $\forall (f,g,h) \in \I^3$, $(fRg$ and $gRh) \Rightarrow fRh$. Let us now give a rigorous definition of the multiplicative Asplund's metric using the space of equivalence  classes $\I^{\LIPtimes}$
\begin{equation}\label{eq:das_formal}
	\forall (f^{\protect \LIPtimes},g^{\protect \LIPtimes}) \in (I^{\protect \LIPtimes})^2 , \quad d_{As}^{\protect \LIPtimes}(f^{\protect \LIPtimes},g^{\protect \LIPtimes}) = d_{As}^{\protect \LIPtimes}(f_1,g_1)
\end{equation}
$d_{As}^{\protect \LIPtimes}(f_1,g_1)$ is defined by eq. \ref{eq:das} between two elements $f_1$ and $g_1$ of the equivalence classes $f^{\protect \LIPtimes}$ and $g^{\protect \LIPtimes}$.

The demonstration of the metric properties are in the appendix (section \ref{sec:appendix}). 

Several examples have shown the interest of using Asplund's metric for pattern matching between a template function $t: D_t \rightarrow \Tcurv$ and the function $f$. For each point $x$ of $D$, the distance $d_{As}^{\LIPtimes} (f_{\left|D_t(x)\right.},t)$ is computed in the neighbourhood $D_t(x)$ centred in $x$, with $f_{\left|D_t(x)\right.}$ being the restriction of $f$ to $D_t(x)$.  Therefore, one can define a map of Asplund's distances \cite{Noyel2015}.

\begin{definition}{\textbf{Map of Asplund's distances}}
Given a grey-level image $f \in \I^*$ and a probe $t \in (\Tcurv^*)^{D_{t}}$, $t>0$, their map of Asplund's distances is:
\begin{equation}\label{eq:map_As}
	As_{t}^{\LIPtimes} f : \left\{
	\begin{array}{ccc}
		\I^* \times (\Tcurv^*)^{D_{t}} &\rightarrow& \left(\Real^{+} \right)^{D}\\
		(f,t) &\rightarrow & As_{t}^{\LIPtimes} f(x) = d_{As}^{\LIPtimes} (f_{\left|D_t(x)\right.},t)\\
	\end{array}
	\right.
\end{equation}
$D_t(x)$ is the neighbourhood associated to $D_t$ centred in $x \in D$.
\end{definition}
One can notice that the template $t$ is acting like a structuring element.

\subsection{Short reminder on Mathematical Morphology}

In this subsection we give a reminder of the basis notions used in Mathematical Morphology (MM) \cite{Matheron1967}.
MM is defined in complete lattices \cite{Serra1982,Heijmans1990,Banon1993,Birkhoff1967}.

\begin{definition}{\textbf{Complete lattice}}
Given a set $\Lscr$ and a binary relation $\leq$ defining a partial order on $\Lscr$, we say that $(\Lscr,\leq)$ is a \textit{partially ordered set} or \textit{poset}. $\Lscr$ is a complete lattice if any non empty subset $\mathscr{X}$ of $\Lscr$ has a supremum (a least upper bound) and an infimum (a greatest lower bound). The infimum and the supremum will be denoted, respectively, by $\wedge \mathscr{X}$ and $\vee \mathscr{X}$. Two elements of the complete lattice $\Lscr$ are important: the least element $O$ and the greatest element $I$.
\end{definition}
 The set of images from $D$ to $[0,M]$, $\overline{\I}=[0,M]^D$, is a complete lattice with the partial order relation $\leq$, by inheritance of the complete lattice structure of $[0,M]$. The least and greatest elements are the constant functions $f_0$ and $f_M$ whose values are equal respectively to $0$ and $M$ for all elements of $D$. The supremum and infimum are respectively, for any $\mathscr{X} \subset \overline{\I}$
\begin{equation}\label{eq:pre_sup_inf}
	\begin{array}{ccc}
		\left(\wedge_{\overline{\I}} \mathscr{X}\right)(x) & = & \wedge_{[0,M]} \left\{ f(x):f \in \mathscr{X}, \> x\in D \right\}\\
		\left(\vee_{\overline{\I}} \mathscr{X}\right)(x)   & = & \vee_{[0,M]}   \left\{ f(x):f \in \mathscr{X}, \> x\in D  \right\}.\\
	\end{array}
\end{equation}
The set of functions $\overline{\Real}^D$ is also a complete lattice with $\overline{\Real} = \Real \cup \{-\infty , +\infty \}$ with the usual order $\leq$, like the set of all subsets of $D$, written $\mathcal{P}(D)$, with the set inclusion $\subset$.

\begin{definition}{\textbf{Erosion, dilation, anti-erosion, anti-dilation} \cite{Banon1993}}
Given $\Lscr_1$ and $\Lscr_2$ two complete lattices, a mapping $\psi \in \Lscr_2^{\Lscr_1}$ is\\
\begin{enumerate}
	\item an erosion iff 				\hspace{2.13em} 	$\forall \mathscr{X} \subset \Lscr_1$, $\psi( \wedge \mathscr{X} ) =  \wedge \psi( \mathscr{X} )$, then we write $\varepsilon = \psi$;
	\item a dilation iff 				\hspace{2.50em} 	$\forall \mathscr{X} \subset \Lscr_1$, $\psi( \vee \mathscr{X} ) =  \vee \psi( \mathscr{X} )$, then we write $\delta = \psi$;
	\item an anti-erosion  iff 	\hspace{0.11em} 	$\forall \mathscr{X} \subset \Lscr_1$, $\psi( \wedge \mathscr{X} ) =  \vee \psi( \mathscr{X} )$, then we write $\varepsilon^a = \psi$;
	\item an anti-dilation iff 	\hspace{0em} 			$\forall \mathscr{X} \subset \Lscr_1$, $\psi( \vee \mathscr{X} ) =  \wedge \psi( \mathscr{X} )$, then we write $\delta^a = \psi$.
\end{enumerate}
As the definitions of these mappings apply even to the empty subset of $\Lscr_1$, we have: $\varepsilon(I)=I$, $\delta(O)=O$, $\varepsilon^a(I)=O$ and $\delta^a(O)=I$.
\label{pre:def_morpho_base}
\end{definition}
Erosions and dilations are increasing mappings: $\forall \mathscr{X}, \mathscr{Y} \subset \Lscr_1$, $\mathscr{X} \leq \mathscr{Y} \Rightarrow \psi(\mathscr{X}) \leq \psi(\mathscr{Y})$ while anti-erosions and anti-dilations are decreasing mappings: $\forall \mathscr{X}, \mathscr{Y} \subset \Lscr_1$, $\mathscr{X} \leq \mathscr{Y} \Rightarrow \psi(\mathscr{Y}) \leq \psi(\mathscr{X})$.

\begin{definition}{\textbf{Structuring element} \cite{Serra1988,Heijmans1990,Angulo2011}}
Let us define a pulse function $i_{x,t} \in \overline{\I}$ of level $t$ at the point $x$: 
\begin{equation}
i_{x,t}(x)=t; \qquad i_{x,t}(y)=0 \> \text{ if } \> x \neq y. 
\label{eq:pulse}
\end{equation}

The function $f$ can be decomposed into the supremum of its pulses $f = \vee \left\{i_{x,f(x)},x \in D \right\}$.
It is easy to define dilations and erosions which are not translation-invariant (in the domain $D$). Let $W$ be a map $\overline{\I} \rightarrow \overline{\I}$ associating to each pulse function $i_{x,t} \in \overline{\I}$ a (functional) ``window'' $W(i_{x,t})$. Then the operator $\delta_W: \overline{\I} \rightarrow \overline{\I}$ defined by:
\begin{equation}
\delta_W(f) = \vee \left\{ W(i_{x,f(x)}) , x \in D \right\}
\end{equation}
is a dilation. When all `windows'' $W(i_{x,f(x)})$ are translation invariant (in $D$), they take the form $W(i_{x,f(x)})=B(x)$ with $B(x)=B_x$ being a structuring element (or structuring function).
\end{definition}
In this case the previously defined dilation $\delta$ and erosion $\varepsilon$, in the same lattice $(\overline{\I},\leq)$, can be simplified:
\begin{equation}
\begin{array}{ccc}
(\delta_B(f))(x) 			&=& \vee 	 \left\{ f(x - h) + B(h), h \in D_B \right\} = (f \oplus B) (x)\\
(\varepsilon_B(f))(x) &=& \wedge \left\{ f(x + h) - B(h), h \in D_B \right\} = (f \ominus B) (x)\\
\end{array}
\label{eq:erode_dilate_funct}
\end{equation}
$D_B \subset D$ is the definition domain of the structuring function $B: D_B \rightarrow \overline{\Tcurv}$. The symbols $\oplus$ and $\ominus$ represent the extension to functions \cite{Serra1982} of Minkowski operations between sets \cite{Minkowski1903,Hadwiger1957}.
Notice: in the case of a flat structuring element with its values equal to zero (i.e. $\forall x \in D_B$, $B(x)=0$), we have $\delta_B(f)(x) = \vee 	 \left\{ f(x - h) , h \in D_B \right\}= \delta_{D_B}(f)(x)$ and $\varepsilon_B(f)(x) = \wedge 	 \left\{ f(x + h) , h \in D_B \right\}= \varepsilon_{D_B}(f)(x)$.


%
%

\section{Map of Aplund's distances and mathematical morphology}

We now link the map of Asplund's distances with Mathematical Morphology.

Given $\overline{\Real}^+=[0,+\infty]$ a complete lattice with the natural order $\leq$, the map of the least upper bounds $\la_B$ between the probe $B \in (\Tcurv^*)^{D_{B}}$ and the function $f \in \overline{\I}$ is defined as:
\begin{equation}
	\la_{B} f : \left\{
	\begin{array}{ccc}
		\overline{\I} \times (\Tcurv^*)^{D_{B}} &\rightarrow& \left(\overline{\Real}^{+} \right)^{D}\\
		(f,B) &\rightarrow & \la_{B} f(x) =  \wedge \left\{\alpha(x), f(x+h) \leq \alpha(x) \LIPtimes B(h), h \in D_B \right\}.\\
	\end{array}
	\right.
	\label{eq:upper_map}
\end{equation}

The map of the greatest lower bounds $\mu_{B}$ between the probe $B \in (\Tcurv^*)^{D_{B}}$ and the function $f \in \overline{\I}$ is defined as:
\begin{equation}
	\mu_{B} f : \left\{
	\begin{array}{ccc}
		\overline{\I} \times (\Tcurv^*)^{D_{B}} &\rightarrow& \left(\overline{\Real}^{+} \right)^{D}\\
		(f,B) &\rightarrow & \mu_{B} f(x) =  \vee \left\{\beta(x), \beta(x) \LIPtimes B(h) \leq f(x+h) , h \in D_B \right\}.\\
	\end{array}
	\right.
	\label{eq:lower_map}
\end{equation}
The two mappings $\la_{B}$ and $\mu_{B}$ are defined between two complete lattices $\Lscr_1 = (\overline{\I} , \leq)$ and $\Lscr_2 = (\left(\overline{\Real}^{+} \right)^{D},\leq)$ with the natural order $\leq$.
Therefore, the least element of $(\Lscr_1 , \leq)$ corresponds to the constant function equal to zero, $O = f_0$ and the greatest element is the constant function equal to $M$, $I = f_M$.

Using the equations \ref{eq:upper_map} and \ref{eq:lower_map} the map of Asplund's distances (eq. \ref{eq:map_As}) can be simplified:
\begin{equation}
	As_B^{\LIPtimes}f = \ln \left( \frac{ \la_{B} f }{ \mu_{B} f }\right), \text{ with } f > 0.
	\label{eq:map_As_la_mu}
\end{equation}

In addition, $\forall x \in D$, $\forall h \in D_B$, $\forall \alpha \in \Real^+$, we have:
\begin{equation}
\begin{array}{lcl}
 \alpha(x) \LIPtimes B(h) \geq f(x+h) & \Leftrightarrow & M - M \left( 1 - B(h)/M \right)^{\alpha(x)} \geq f(x+h), \> \text{ (from eq. \ref{eq:LIPtimes})}\\
																	 & \Leftrightarrow & \alpha(x) \geq \frac{ \ln{\left( 1 - \frac{f(x+h)}{M} \right)} }{ \ln{\left( 1 - \frac{B(h)}{M} \right)} }, \> \text{because } \left( 1 - \frac{B(h)}{M} \right) \in ]0,1[.
\end{array}
\label{eq:dem_flat_se}
\end{equation}

We assume that $\widetilde{f}= \ln{\left( 1 - f/M \right)}$. Using equation \ref{eq:dem_flat_se}, equation \ref{eq:upper_map} becomes:
\begin{equation}
\la_{B} f = \wedge \left\{\alpha(x), \alpha(x) \geq \frac{ \widetilde{f}(x+h) }{ \widetilde{B}(h) }, h \in D_B \right\} 
	= \vee \left\{ \frac{ \widetilde{f}(x+h) }{ \widetilde{B}(h) }, h \in D_B \right\}.
\label{eq:upper_map_2}
\end{equation}
In a similar way:
\begin{equation}
\mu_{B} f = \vee \left\{\beta(x), \beta(x) \leq \frac{ \widetilde{f}(x+h) }{ \widetilde{B}(h) }, h \in D_B \right\}
= \wedge \left\{ \frac{ \widetilde{f}(x+h) }{ \widetilde{B}(h) }, h \in D_B \right\}.
\label{eq:lower_map_2}
\end{equation}

\subsection{Case of a flat structuring element}

In the case of a flat structuring element $B=B_0 \in \Tcurv^*$ ($\forall x \in D_B$, $B(x) = B_0$), the equations \ref{eq:upper_map_2} and \ref{eq:lower_map_2} can be simplified. 
\begin{equation}
\begin{array}{ccl}
	\la_{B_0} f &=& \frac{1}{\widetilde{B}_0} \wedge \left\{ \widetilde{f}(x+h) ,  h \in D_B \right\}, \quad \text{ because } \widetilde{B}_0 < 0\\
						&=& \frac{1}{\widetilde{B}_0} \ln{ \left( 1 - \frac{ \vee \{ f(x-h) ,  -h \in D_B \} }{ M }\right) }\\
						&=& \frac{1}{\widetilde{B}_0} \ln{ \left( 1 - \frac{ \delta_{\widecheck{D}_B} f }{ M }\right) } \label{eq:upper_map_flat_se}
\end{array}
\end{equation}
Notice: the infimum $\wedge$ is changed into a supremum $\vee$ because the function $\widetilde{f}: x \rightarrow \ln(1-x/M)$ is a continuous decreasing mapping.
The reflected (or transposed) domain $\widecheck{D}_B$ is $\widecheck{D}_B = \{-h, h\in D_B \}$ and the reflected structuring function $\widecheck{B}$ is defined by the reflection of its definition domain $\forall x \in \widecheck{D}_B$, $\widecheck{B}(x)=B(-x)$ \cite{Soille2003}.
Similarly: 
\begin{equation}
\begin{array}{ccl}
	\mu_{B_0} f &=& \frac{1}{\widetilde{B}_0} \vee \left\{ \widetilde{f}(x+h) ,  h \in D_B \right\}, \quad \text{ because } \widetilde{B}_0 < 0\\
						&=& \frac{1}{\widetilde{B}_0} \ln{ \left( 1 - \frac{ \wedge \{ f(x+h) ,  h \in D_B \} }{ M }\right) }\\
						&=& \frac{1}{\widetilde{B}_0} \ln{ \left( 1 - \frac{ \varepsilon_{D_B} f }{ M }\right) }.
\end{array}
\label{eq:lower_map_flat_se}
\end{equation}

With the equations \ref{eq:upper_map_flat_se} and \ref{eq:lower_map_flat_se}, the map of Asplund's distances (eq. \ref{eq:map_As_la_mu}) becomes:
\begin{equation}
	As_{B_0}^{\LIPtimes}f = \ln{ \left[ \frac{ \ln{ \left( 1 - \frac{ \delta_{\widecheck{D}_B} f }{ M }\right) } }{ \ln{ \left( 1 - \frac{ \varepsilon_{D_B} f }{ M }\right) } }\right] }, \> \text{ with } f > 0.
	\label{eq:map_As_la_mu_flat_se}
\end{equation}

This important result shows that, with a flat probe, the map of Apl\"und's distances can be computed using logarithms and operations of morphological erosion and dilation of an image. From an implementation point of view, the programming of the map of Apl\"und's distances becomes easier, because the majority of image processing libraries contains morphological operations.

Notice: by replacing the dilation and erosion by rank-filters \cite{Serra1988} one can compute the map of Asplund's distances with a tolerance \cite{Jourlin2014,Noyel2015}.

\subsection{General case: a structuring function}

Using a general structuring function in the equations \ref{eq:upper_map_2} and \ref{eq:lower_map_2}, the map of Asplund's distances is expressed as:
\begin{equation}
	As_{B}^{\LIPtimes}f = \ln \left( \frac{ \la_{B} f }{ \mu_{B} f }\right) = 
	\ln{ \left( \frac{ \vee \left\{ \frac{ \widetilde{f}(x+h) }{ \widetilde{B}(h) }, h \in D_B \right\} }{ \wedge \left\{ \frac{ \widetilde{f}(x+h) }{ \widetilde{B}(h) }, h \in D_B \right\} }\right) }, \> \text{ with } f > 0.
	\label{eq:map_As_la_mu_general_se}
\end{equation}

Let us study the properties of mappings $\la_{B}$, $\mu_{B} \in \Lscr_{2}^{\Lscr_1}$, $\forall f, g \in \overline{\I}$
\begin{equation}
\begin{array}{ccl}
	\la_{B} ( f \vee g ) &=& \vee \left\{ \frac{ \widetilde{f \vee g }(x+h) }{ \widetilde{B}(h) }, h \in D_B \right\}\\
	&=& \vee \left\{ \frac{ \widetilde{f}(x+h) \wedge \widetilde{g}(x+h) }{ \widetilde{B}(h) }, h \in D_B \right\}, \> \text{because $\widetilde{f}$ is decreasing}\\
	&=& \vee \left\{ \frac{ \widetilde{f}(x+h)}{ \widetilde{B}(h) } \vee \frac{ \widetilde{g}(x+h) }{ \widetilde{B}(h) }, h \in D_B \right\}, \> \text{because } \widetilde{B}(h) <0\\
	&=& \left[ \vee \left\{ \frac{ \widetilde{f}(x+h) }{ \widetilde{B}(h) }, h \in D_B \right\} \right] \vee \left[ \vee \left\{ \frac{ \widetilde{g}(x+h) }{ \widetilde{B}(h) }, h \in D_B \right\} \right]\\
	&=& \la_{B} ( f ) \vee \la_{B} ( g ).
\end{array}
\label{eq:upper_map_gen_se_dilation}
\end{equation}
According to definition \ref{pre:def_morpho_base}, 2 (p. \pageref{pre:def_morpho_base}), $\la_{B}$ is a dilation.
In addition, 
\begin{equation}
\la_B(O) = \la_B(f_0) = \wedge \left\{\alpha(x), \alpha(x) \geq \frac{ \widetilde{0}(x+h) }{ \widetilde{B}(h) }, h \in D_B \right\} = 0 = O.
\label{eq:smallest_elt_la}
\end{equation}
Similarly, we have:
\begin{equation}
\begin{array}{ccl}
	\mu_{B} ( f \wedge g ) 
	&=& \wedge \left\{ \frac{ \widetilde{f \wedge g }(x+h) }{ \widetilde{B}(h) }, h \in D_B \right\}\\
	&=& \wedge \left\{ \frac{ \widetilde{f}(x+h) \vee \widetilde{g}(x+h) }{ \widetilde{B}(h) }, h \in D_B \right\}, \> \text{because $\widetilde{f}$ is decreasing}\\
	&=& \left[ \wedge \left\{ \frac{ \widetilde{f}(x+h) }{ \widetilde{B}(h) }, h \in D_B \right\} \right] \wedge \left[ \wedge \left\{ \frac{ \widetilde{g}(x+h) }{ \widetilde{B}(h) }, h \in D_B \right\} \right], \> \text{because } \widetilde{B}(h) <0\\
	&=& \mu_{B} ( f ) \wedge \mu_{B} ( g ).
\end{array}
\label{eq:lower_map_gen_se_erosion}
\end{equation}
According to definition \ref{pre:def_morpho_base}, 1  (p. \pageref{pre:def_morpho_base}), $\mu_{B}$ is an erosion.
In addition, 
\begin{equation}
\mu_B(I) = \mu_B(f_M) = \vee \left\{\beta(x), \beta(x) \leq \frac{ \widetilde{M}(x+h) }{ \widetilde{B}(h) }, h \in D_B \right\} = + \infty = I.
\label{eq:greatest_elt_mu}
\end{equation}

Therefore, the map of Asplund's distances is the logarithm of the ratio between a dilation and an erosion of the function $f$ by the structuring function $B$. The  map of the least upper bounds $\la_B$ is a dilation and the map of the greatest lower bounds $\mu_B$ is an erosion. The two maps are defined from the lattice $(\Lscr_1 = \overline{\I} , \leq)$ and the lattice $(\Lscr_2 = (\overline{\Real}^{+} )^{D},\leq)$ with their respective natural orders.

%
%

\section{Illustration}

\begin{figure}
\begin{center}
\begin{tabular}{ccc}
  \includegraphics[width=0.33\columnwidth]{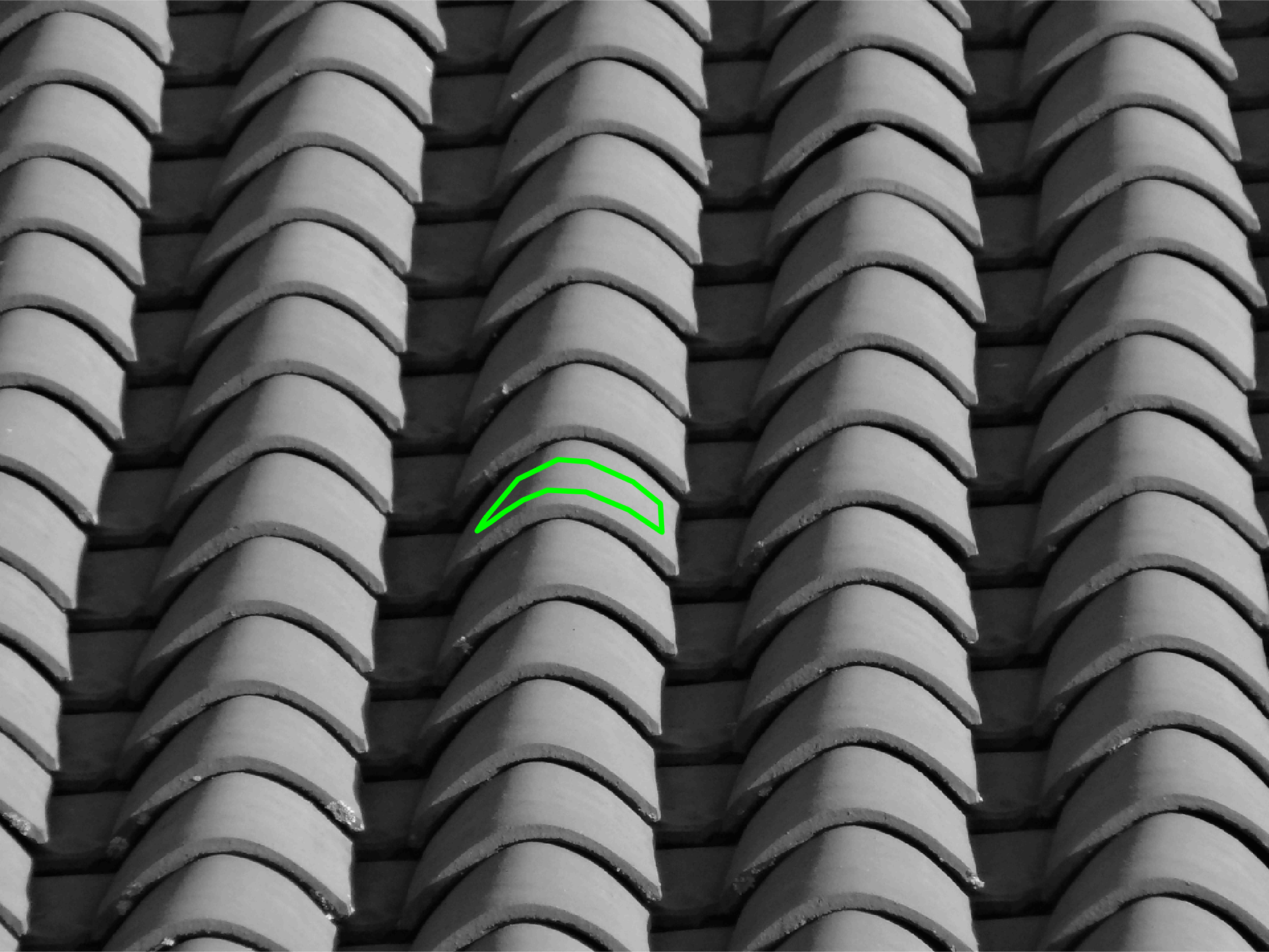}&
	\includegraphics[width=0.33\columnwidth]{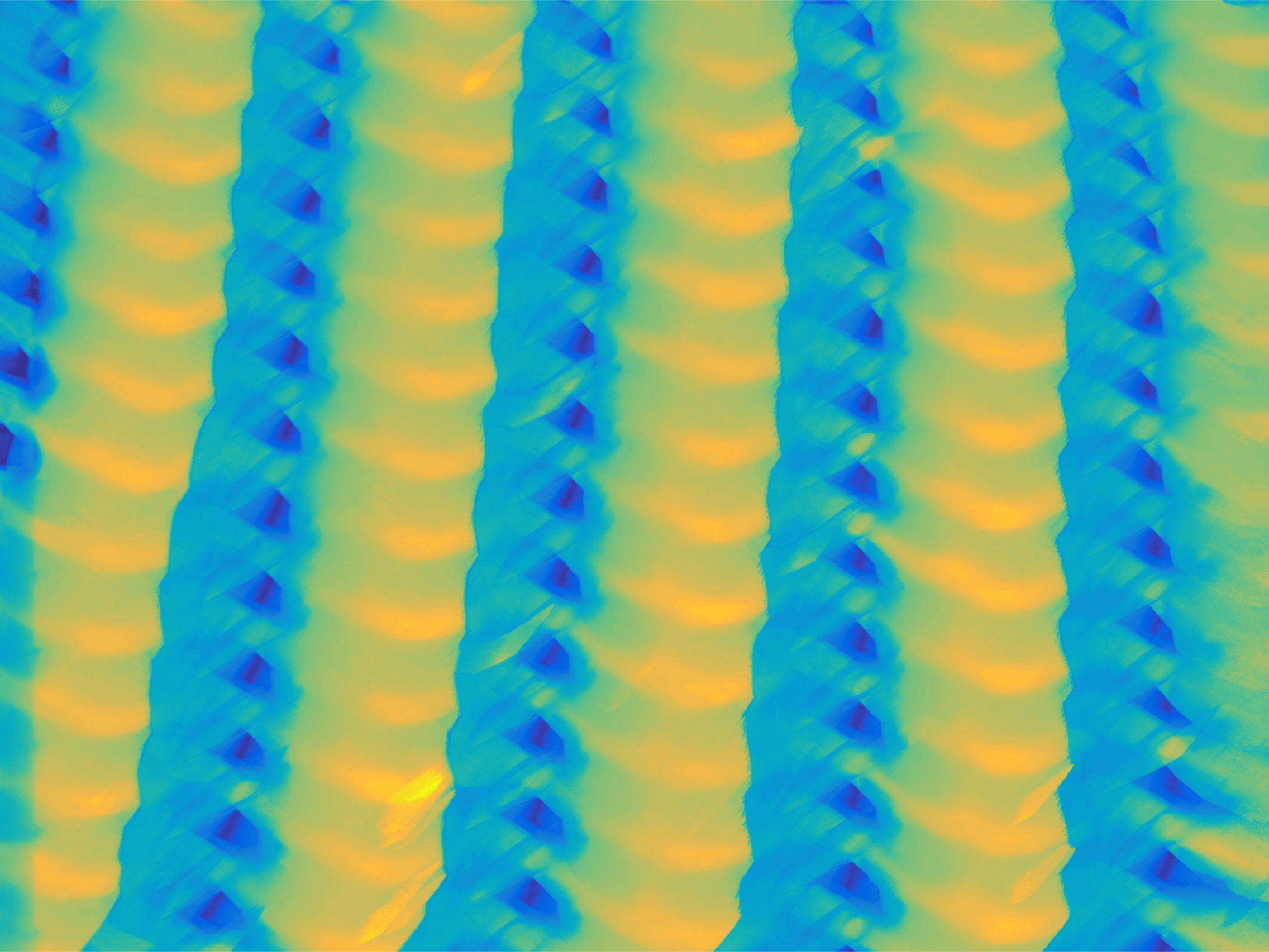}&
	\includegraphics[width=0.33\columnwidth]{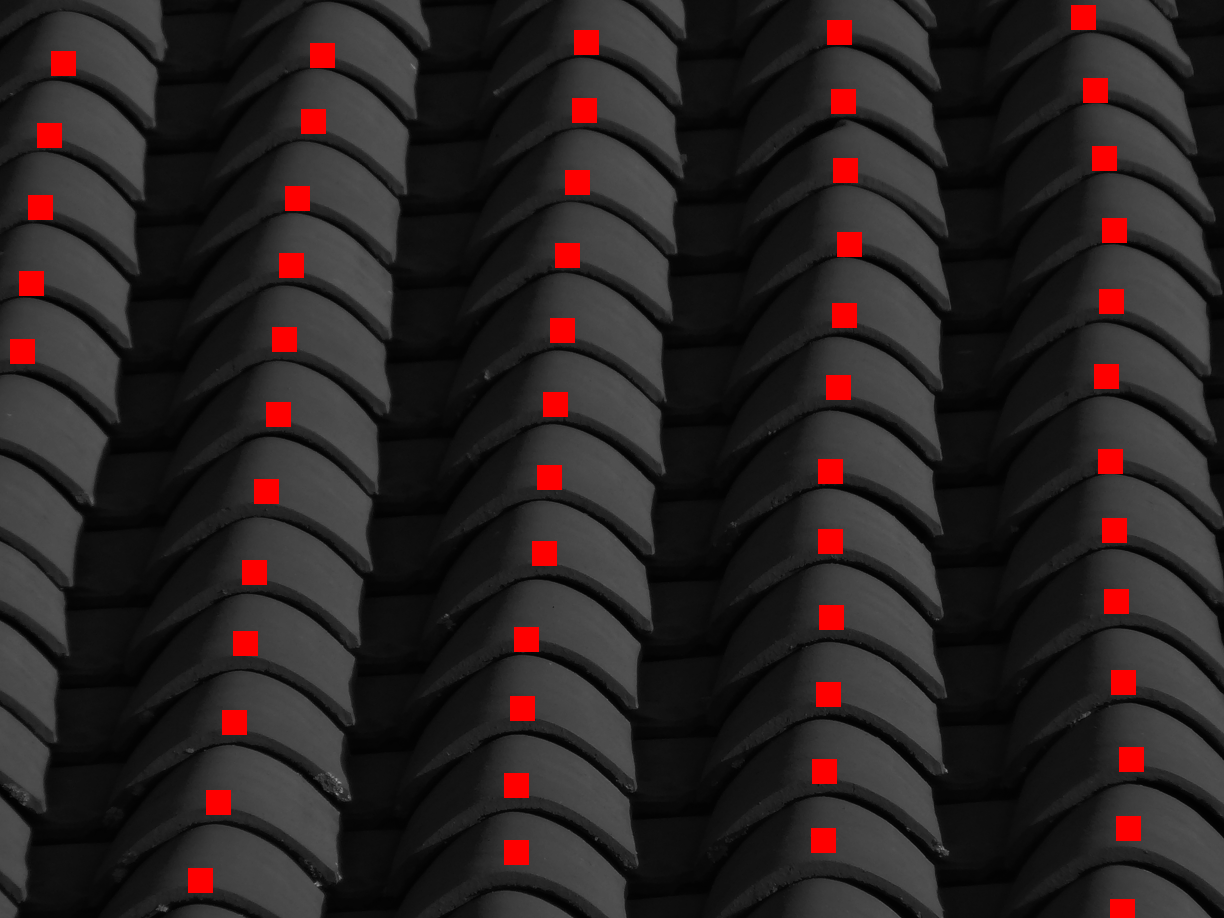}\\
(a) Image $f$ and probe $B$&
(b) Map $As_{B,p}^{\LIPtimes} f^d$&
(c) Detected tiles in $f^d$\\
\end{tabular}
\end{center}
\caption{Detection of tiles using the map of Asplund's distances $As_{B,p}^{\protect \LIPtimes} f^d$ with a tolerance $p = 30 \%$. (a) The probe $B$ (in green) is extracted in the image $f$. (b) The minima (in blue) of the map of distances, $As_{B,p}^{\protect \LIPtimes} f^d$, between the probe and the darken image $f^d$ are extracted with a threshold of 0.7. (c) Location of the detected tiles (red dots) in the darken image $f^d$.}
\label{fig:tiles_detection}
\end{figure}

In figure \ref{fig:tiles_detection} (a), we extract a tile (i.e. the probe or the structuring function) in an image $f$ and we look for the similar ones in a darken image, $f^d$, by means of a LIP multiplication of 0.3. Physically, it corresponds to an object with a stronger light absorption. Importantly, the probe has a non convex domain shape and is not flat. We compute the map of Asplund's distances between the probe $B$ and the image $f^d$ with a tolerance, $As_{B,p}^{\LIPtimes} f^d$, as introduced in \cite{Jourlin2014,Noyel2015}. This  metric, robust to noise, is computed by discarding $p = 30 \%$ of the points which are the closest to the least upper bounds and to the greatest lower bounds. The tiles are located at the local minima of the distance map which are extracted by a threshold of 0.7 (fig. \ref{fig:tiles_detection} (b)). The tiles similar to the probe, according to the Asplund's distance, have been correctly detected (fig. \ref{fig:tiles_detection} (c)). Notice that the domain of the probe is slightly smaller than the domain of the tiles.

%
%

\section{Conclusion}

In the current paper, we have shown that the map of Asplund's distances between a probe and a function using the LIP multiplication is linked with morphological operations. The probe corresponds to a structuring function and the map of Asplund's distances is the logarithm of the ratio between a dilation and an erosion of the function by the structuring function into the lattice of positive functions $((\overline{\Real}^{+} )^{D},\leq)$. The dilation is the map of the least upper bounds $\la_B f$, between the function $f$ and the probe $B$, while the erosion is the map of the greatest lower bounds $\mu_B f$. 

The dilation and the erosion are mappings between the complete lattices of the images $(\overline{\I} , \leq)$ and the lattice $((\overline{\Real}^{+} )^{D},\leq)$ with the natural order. When using a flat structuring element, the expression of the map of Asplund's distances can be simplified with a dilation and an erosion of the image into the same lattice of the images $(\overline{\I} , \leq)$. An example of pattern matching has been presented with a non-flat structuring function.  

The obtained results set the pattern matching approach by Asplund's distances in the well established framework of Mathematical Morphology. The current reasoning can be extended to the double-sided probing by Asplund's distances for colour and multivariate images using the LIP multiplicative or the LIP additive framework \cite{Noyel2015,Jourlin2016_chap3,Noyel2016}. This will be presented in a coming paper. 

%
%

\section{Appendix}
\label{sec:appendix}

Le us demonstrate that the Apl\"und's metric $d_{As}^{\LIPtimes}$ is a metric in the space of equivalence classes $\I^{\LIPtimes}$. In order to be a metric on $(\Ieq \times \Ieq) \rightarrow \Real^+$, $d_{As}^{\LIPtimes}$ must satisfy the four following properties:
\begin{enumerate}

	\item (positivity): 
	$\forall \ft \neq \gt \in \Ieq$, $\forall x \in D$, $\la \LIPtimes \gt(x) > \mu \LIPtimes \gt(x)$ (def. \ref{def:asplund_metric}, p. \pageref{def:asplund_metric}), because  $\gt>0$
	
	$\Rightarrow \la > \mu$ because $\Ieq$ is an ordered set with the order $\leq$
	
	$\Rightarrow \forall \ft \neq \gt \in \Ieq$, $d_{As}^{\LIPtimes}(\ft,\gt)>0$. 
	
	
	\item (Axiom of separation): 
\begin{equation}	
\begin{array}{lll}
\left.
\begin{array}{l}
	d_{As}^{\LIPtimes}(\ft,\gt) = 0 \Rightarrow \la = \mu\\
	(\text{def. \ref{def:asplund_metric}}) \Rightarrow \la \LIPtimes \gt \geq \ft \geq \mu \LIPtimes \gt
\end{array}
\right\} &
\Rightarrow \la \LIPtimes \gt = \ft & 
\Rightarrow \ft = \gt \text{ in } \Ieq \\
\end{array}
\label{eq:dem_dAs_axiom_separ_1}
\end{equation}	
Reciprocally:
\begin{equation}	
\begin{array}{ll}
\left.
\begin{array}{l}
	\forall \ft, \gt \in \Ieq, \ft = \gt \Rightarrow \la \LIPtimes \gt = \ft\\
	(\text{def. \ref{def:asplund_metric}}) \la \LIPtimes \gt \geq \ft \geq \mu \LIPtimes \gt
\end{array}
\right\} &
\Rightarrow  \la \LIPtimes \gt =\ft = \mu \LIPtimes \gt\\
\\
\Rightarrow \la = \mu 
\Rightarrow d_{As}^{\LIPtimes}(\ft,\gt) = 0
\end{array}
\label{eq:dem_dAs_axiom_separ_2}
\end{equation}	

Eq. \ref{eq:dem_dAs_axiom_separ_1} and \ref{eq:dem_dAs_axiom_separ_2} $\Rightarrow$ $\left\{\forall \ft, \gt \in \Ieq \right.$, $\left. d_{As}^{\LIPtimes}(\ft,\gt) = 0 \Leftrightarrow \ft = \gt \right\}$.

	\item (Triangle inequality): 
Let us define: $d_{As}^{\LIPtimes}(\ft,\gt) = \ln(\la_1/\mu_1)$, $d_{As}^{\LIPtimes}(\gt,\htt) = \ln(\la_2/\mu_2)$ and $d_{As}^{\LIPtimes}(\ft,\htt) = \ln(\la_3/\mu_3)$.
We have 
\begin{equation} \label{eq:dem_ineq_1}
	d_{As}^{\LIPtimes}(\ft,\gt) + d_{As}^{\LIPtimes}(\gt,\htt) = \ln\left( \frac{\la_1 \la_2}{\mu_1 \mu_2}\right)
\end{equation}

\begin{equation*}	
\begin{array}{lcl}
\text{Def. \ref{def:asplund_metric} } &\Rightarrow& \left\{
\begin{array}{l}
 \la_1 = \inf\left\{k_1: \forall x, k_1 \LIPtimes \gt(x) \geq \ft(x)\right\}\\
 \la_2 = \inf\left\{k_2: \forall x, k_2 \LIPtimes \htt_j(x) \geq \gt(x)\right\}\\
 \la_3 = \inf\left\{k_3: \forall x, k_3 \LIPtimes \htt_j(x) \geq \ft(x)\right\}\\
\end{array}\right.\\
\end{array}
\end{equation*}

\begin{equation}
\begin{array}{llll}
\Rightarrow& \la_1 \la_2 & \leq& \inf_{k_1} \left\{ \inf_{k_2} \left\{ \forall x, k_1 \LIPtimes ( k_2 \LIPtimes \htt_j(x)) \geq k_1 \LIPtimes \gt(x) \right\} \geq \ft(x)  \right\}\\
&&\leq& \inf \left\{ k': \forall x, k' \LIPtimes \htt_j(x) \geq \ft(x)\right\} \text{, with } k' = k_1 \times k_2\\
\Rightarrow& \la_1 \la_2 & \leq & \la_3 \text{,  with } \la_1, \la_2, \la_3 > 0\\
\end{array}
\label{eq:dem_ineq_2}
\end{equation}

In the same way:
\begin{equation} 
\mu_1 \mu_2 \geq \mu_3 \text{,  with } \mu_1, \mu_2, \mu_3 > 0
\label{eq:dem_ineq_3}
\end{equation}

Eq. \ref{eq:dem_ineq_1}, \ref{eq:dem_ineq_2} \ref{eq:dem_ineq_3} $\Rightarrow$ $\frac{\la_1 \la_2}{\mu_1 \mu_2} \geq \frac{\la_3}{\mu_3}$

$\Rightarrow \forall \ft, \gt, \htt \in \Ieq, d_{As}^{\LIPtimes}(\ft,\htt) \leq d_{As}^{\LIPtimes}(\ft,\gt) + d_{As}^{\LIPtimes}(\gt,\htt)$.

	\item (Axiom of symmetry): 
Let us define: $d_{As}^{\LIPtimes}(\ft,\gt) = \ln(\la_1/\mu_1)$, $d_{As}^{\LIPtimes}(\gt,\ft) = \ln(\la_2/\mu_2)$.

Def. \ref{def:asplund_metric} $\Rightarrow$
$\la_1 = \inf \left\{ k: \forall x, \gt \geq \frac{1}{k} \LIPtimes \ft \right\}$, because $k>0$

$\Rightarrow \frac{1}{\la_1} = \sup \left\{ k': \forall x, \gt \geq k' \LIPtimes \ft\right\}$
$\Rightarrow \frac{1}{\la_1} = \mu_2$.

In the same way, we have $\frac{1}{\mu_1} = \la_2$.

Therefore, $\forall \ft, \gt \in \Ieq$, $d_{As}^{\LIPtimes}(\ft,\gt) = \ln(\la_1/\mu_1) = \ln(\la_2/\mu_2) = d_{As}^{\LIPtimes}(\gt,\ft)$.

\end{enumerate}

%
%

\bibliographystyle{splncs03}
\bibliography{refs}

\end{document}